\documentclass{article}
\usepackage{ijcai19}

\usepackage{times}
\usepackage{soul}
\usepackage{url}
\usepackage[hidelinks]{hyperref}
\usepackage[utf8]{inputenc}
\usepackage[small]{caption}
\usepackage{booktabs}
\usepackage{graphicx}
\usepackage{amsmath,amssymb}
\usepackage{makecell}
\usepackage{multirow}

\title{Learning from Observations Using a Single Video Demonstration and Human Feedback}

\author{
Sunil Gandhi$^1$ \and 
Tim Oates$^1$ \and
Tinoosh Mohsenin$^1$ \and
Nicholas Waytowich$^2$
\affiliations
$^1$University of Maryland Baltimore County \\
$^2$US Army Research Laboratory 
\emails
\{sunilga1,oates,tinoosh\}@umbc.edu,
nicholas.r.waytowich.civ@mail.mil
}

\begin{document}

\maketitle

\begin{abstract}
In this paper, we present a method for learning from video demonstrations by using human feedback to construct a mapping between the standard representation of the agent and the visual representation of the demonstration. In this way, we leverage the advantages of both these representations, i.e., we learn the policy using standard state representations, but are able to specify the expected behavior using video demonstration. We train an autonomous agent using a single video demonstration and use human feedback (using numerical similarity rating) to map the standard representation to the visual representation with a neural network. We show the effectiveness of our method by teaching a hopper agent in the MuJoCo to perform a backflip using a single video demonstration generated in MuJoCo as well as from a real-world YouTube video of a person performing a backflip. Additionally, we show that our method can transfer to new tasks, such as hopping, with very little human feedback. 
\end{abstract}

\section{Introduction}

\par Recent advances in reinforcement learning and robotics have enabled usage of autonomous agents in numerous consumer applications. These systems offer the potential for autonomous agents to be used in a variety of applications like elder care and for performing household chores. However easy interactive methods for teaching new skills to robotic agents are lagging. As robots are becoming more prevalent, the need for easy-to-use methods for non-expert users to teach new skills is increasing. 

\par A common way to train a reinforcement learning agent is by optimizing the policy to maximize a well-specified reward function. But, even for experts, designing reward functions is a complex and time-consuming process, and for some tasks, they are too difficult to be specified by hand \cite{Christiano}. Consequently, other approaches for teaching new skills to autonomous agents have been explored. A common method for teaching new skills is through example demonstrations. Often, these demonstrations are collected in the form of lower dimensional representations like angles of joints of a robotic arm  (often referred to as the standard representation). But, collection of demonstrations using standard representation is a difficult process and requires expertise. Another approach proposed by \cite{Sermanet} learns the expected behavior by observing the task being performed in a video recording. This method of providing expected behavior is simple, intuitive and can leverage a large number of videos on the web. However, it requires multiple videos of the same task from multiple viewpoints. Also, this method does not allow for any corrective feedback to improve the performance of the agent. There are other methods like the ones proposed in \cite{Christiano}, \cite{Warnell} that learn the task using human feedback. However, these methods do not use the simple and intuitive method of showing the expected behavior through videos. 

\par We believe that a combination of these approaches could result in a powerful method for teaching autonomous agent to perform a task. We leverage the lower dimensional standard representation to learn the policy of a reinforcement learning agent. Usage of standard representation simplifies the optimization of the reinforcement learning algorithm as it does not have to deal with the complexities of understanding the video. But, like \cite{Sermanet}, we use a simple and intuitive method of using video recording to demonstrate the task. Finally, we utilize human feedback to map the standard and visual representations. Mapping standard and visual representations is a difficult problem for autonomous agents especially if the number of example behaviors is limited. However, this can be easily performed by humans. In this paper, we ask human participants to rate the similarity between behaviors produced by the learning agent alongside example video clips of the desired behavior.  We show that human feedback enables us to learn complex behaviors from a single video demonstration. Usage of a single demonstration is important as demonstrating the same task multiple times is often more difficult than providing ratings indicating the similarity between the agents and expected behavior. 

\par To summarize, the main contributions of this paper are three-fold:
First, we present a novel system to teach new skills to an autonomous agent using a single video demonstration and human feedback. The agent learns the policy using a standard representation and uses human feedback to map the standard representation to the visual representation. Next, we show effectiveness of our method by teaching an agent to perform backflips in the MuJoCo physics simulator \cite{todorov2012mujoco} from a demonstration video generated in MuJoCo as well as from a real-world video clip from YouTube. Finally, we show the generalization of our method, trained originally to perform a backflip, to replicate other demonstrations like hopping.

\section{Related Work}
\label{sec:RelatedWork}

\par The goal of our system is to create easy interactive ways to teach new skills to autonomous agents. Our approach to this problem is to learn a policy using standard representation, but specify the expected behavior using video demonstration of the task. The mapping between both these representations is learned through human feedback. Even though most of these components have been studied separately to the best of our knowledge there is no method that combines human feedback \cite{Warnell}, reinforcement learning over standard representation \cite{Christiano} and learning from video observations \cite{Sermanet}. Our framework consists of a novel combination of various aspects from these methods that enables us to learn from a single video demonstration and iteratively improve task performance using human feedback. 

\par \textbf{Reinforcement Learning (RL):} RL, and it's deep learning variant Deep-RL, has shown success in a vast array of tasks including playing Atari games \cite{mnih2013playing}, the game of Go \cite{44806} and performing robotic maneuvers \cite{schulman2015trust}. However, defining the reward function for reinforcement learning is often a difficult problem even for experts. Consequently, several alternatives to explicitly defining reward functions have been explored in the literature. This includes methods like behavior cloning \cite{billard2013robot}, \cite{abbeel2004apprenticeship} and Inverse reinforcement learning (IRL) \cite{ng2000algorithms}, \cite{abbeel2004apprenticeship} that learn robotic skills from expert demonstrations. Both these methods require expert demonstrations in the form of state-action pairs. They require the state representation used to demonstrate the expected behavior and to train the reinforcement learning policy to be the same. Two common ways of achieving this are through the use of kinesthetic demonstrations \cite{calinon2007learning} or teleoperation \cite{pastor2009learning}. The advantage of both these methods is that they can collect expert demonstrations in form of lower level state representation like angles of joints of a robotic arm. This simplifies the training of autonomous agents as they don't have to deal with complexities of understanding the visual representation. But collecting demonstrations using kinesthetic demonstration and teleoperation not only require human expertise but can also be expensive to collect.

\par \textbf{Learning from observation} Recently, several methods that learn robotic manipulations from video demonstrations have been proposed \cite{stadie2017third}, \cite{Sermanet}, \cite{sermanet2016unsupervised}, \cite{lee2017learning}, \cite{Liu2018}. Unlike kinesthetic demonstration and teleoperation, using a video to specify the expected behavior is simple and does not require expertise (as the video does not contain actions but only sequences of observations). However, training these methods typically requires a large number of demonstration videos. For example, \cite{Sermanet} uses multiple videos of the task from multiple viewpoints for training. In contrast, we teach our agent to perform backflipping task using only a single video demonstration. Also, unlike our method, these methods are fully automated and do not allow for human feedback. Another recent method, \cite{2018-TOG-SFV} learns to perform acrobatic maneuvers from videos by estimating the pose of the actor, creating a reference trajectory and using it to train an RL agent. \cite{2018-TOG-SFV} requires the use of a pre-trained pose estimator which has the benefit of being fully automated but is restricted to the domain limitations of the pose estimator. In contrast, our method, works on tasks where the kinematics between the agent and demonstrator are quite different. This enables our approach to be useful for non-humanoid robots as well as tasks to which pose estimation may not be available. The learning task in Section \ref{sec:syn_demo} is a good example of this.

\par \textbf{Learning from human feedback:} There is a large amount of literature that allow autonomous agents to learn through human feedback \cite{akrour2011preference}, \cite{wirth2016model}. One of the popular methods for training using human feedback is TAMER (Training an Agent Manually via Evaluative Reinforcement) \cite{knox2008tamer}. In TAMER, the user observes the agents behavior and gives scalar feedback indicating whether current action the agent is executing is good or bad. The agent uses this feedback to learn a better policy \cite{Warnell}. Another recent work by \cite{Christiano} learns expected behavior from human preferences. \cite{Christiano} asks humans to compare two short video clips and ask which one is preferred. It uses this feedback to train a reward predictor that gives higher reward to the preferred behavior. Although these methods utilize human feedback, none of these methods learn from video demonstrations.

\section{Architecture}
\label{sec:arch}

\begin{figure}[tb]
\begin{center}
\includegraphics[width=0.95\columnwidth]{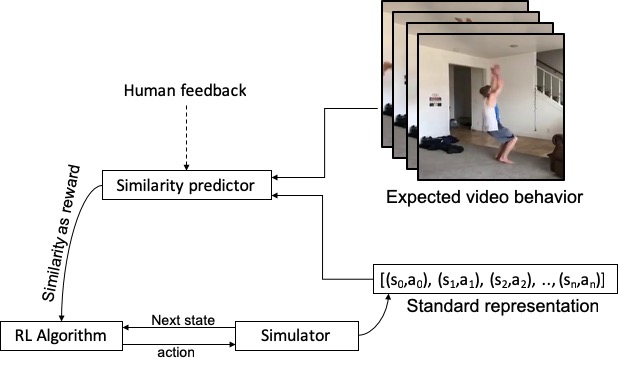}
\end{center}
\caption{Block diagram of our approach. The Reinforcement learning algorithm (TRPO) interacts with the simulator to generate trajectories in the standard representation. These trajectories are compared with a video demonstration using a similarity predictor which is learned using human feedback. The similarity predictor's learned rating is used as a reward signal for the RL algorithm. Both the Similarity predictor and RL algorithm are trained simultaneously.}
\label{fig:architecture}
\end{figure}

\par The goal of our framework is to teach a reinforcement learning agent to replicate the behavior shown in a video demonstration. To train this agent, we use a single video demonstration of the task and human feedback that indicates the similarity between the agent's behavior and the video demonstration.  

\par The overall architecture of our system is illustrated in Figure \ref{fig:architecture}. Our approach is derived from the work of \cite{Christiano} that asks human participants to compare two short video clips and collects there preferences. We modify this approach to solve learning from observation from videos. To achieve this goal, we use human feedback to learn a mapping betweem visual and standard representation. Consequently, instead of asking for preferred behavior, we ask humans to provide a similarity rating between a small clip from the video demonstration and a clip of the agents behavior. Also, it's important to note that unlike \cite{Christiano}, where learning is performed entirely from standard representations, our method is able to learn from both the standard representation as well as from a visual representation due to our similarity network that is described in more detail below. 

\par To precisely define the problem statement, consider a video demonstration with $n$ frames given by $V = (v_0,v_1,...v_{n-1})$. Let the equivalent behavior in the simulator be given by the standard representation $O = (O_0, O_1,...O_{n-1})$, where $O_i$ is the concatenation of observed state vector $S_i$ and action vector $A_i$ at time $t=i$. Given the video demonstration $V$, our goal is to find the corresponding $O$. We formulate this problem as the maximization of  equation \ref{eqn:archeq}, where $n$ is the number of frames, and $S$ is a similarity function that compares the behavior in the video demonstration with the one performed by the learning agent. 
\small
\begin{equation}
    maximize \sum_{t=0}^{n-1} S(v_t,o_t)
    \label{eqn:archeq}
\end{equation}
\normalsize
\par Notice that the only known variable is the single video demonstration $V$. We learn the similarity function $S$ and use it to optimize for $O$ such that equation \ref{eqn:archeq} is maximized. This maximization is done using the following three processes. First, the similarity function $S$ is learned by training a deep neural network using human feedback collected in the form of similarity ratings between the demonstrated clip and the corresponding behavior of the learning agent. This enables the comparison between any trajectory to the expected behavior provided by the demonstration video. Second, the comparison between the agent's behavior and the expected behavior using the similarity network is used as a reward for optimizing a reinforcement learning algorithm. During optimization, the RL agent generates trajectories $o$ that are similar to the video demonstration with respect to similarity function $S$. Finally, segments of the trajectories generated during the optimization of the reinforcement learning algorithm are provided to human for feedback. These three processes, i.e., training of the similarity network, optimization using reinforcement learning  and collecting human feedback run simultaneously and asynchronously. Under this framework, any traditional reinforcement learning algorithms can be used in principle, however, we use trust region policy optimization (TRPO) as it works well when the reward function is not stationary \cite{schulman2015trust}. 

\begin{figure}[htb]
\begin{center}
\includegraphics[width=0.95\columnwidth]{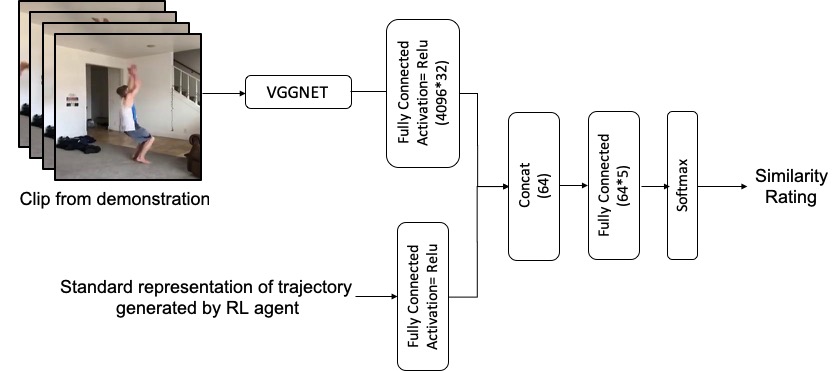}
\end{center}
\caption{Architecture of the similarity prediction network. This network compares a small trajectory segment from the agent to a clip from the video representation and returns a rating between 1 to 5 (1 being not similar and 5 being very similar).}
\label{fig:similarity_network}
\end{figure}

\par To learn the similarity predictor $S$, we use human feedback to provide ratings of how similar the agent is to the video. We show two short clips to the human. One clip is from the video demonstration and another is the corresponding imitation by the RL agent. The video demonstration and imitation clips are about 0.3 seconds long. The human provides feedback as a rating indicating the similarity between both clips ranging from $1$ to $5$, where a rating of $1$ indicates that behaviors in both clips are not at all similar, and $5$ indicates that they are very similar. We collect the feedback in two stages. First, during a pre-training stage, feedback is collected with the agent performing random actions. Later during training, we use trajectories generated by the RL agent for getting human feedback to further refine the similarity predictor. 

\par The similarity predictor is a neural network trained using human feedback to predict the similarity between behaviors in a video clip and behaviors from the RL agent in the standard representation. The architecture of this similarity predictor network is shown in Figure \ref{fig:similarity_network}. The similarity predictor comprises of two branches. One branch extracts features from the frames of the video demonstration clip $v$ and the other extracts features from the standard representation $o$. We use vectors from last fully connected layer of pre-trained VGGNet \cite{simonyan2014very} to extract features from the video frames. For feature extraction of the standard representation, we use two fully connected layers with RELU activation. Both features from standard and visual representation are then concatenated and forwarded through an additional fully connected layer as well as a softmax layer. We train the network to predict human similarity ratings for each frame and observation pair. We treat the prediction of rating as a classification problem with $5$ classes. 
The network is trained with stochastic gradient descent (SGD) with a cross-entropy loss function. Although the basic architecture remains the same, we experiment with several variations and discuss there effect on accuracy in section \ref{sec:Experiments}.

\section{Experiments}
\label{sec:Experiments}

\par We show the effectiveness of our method by teaching an agent to backflip in the MuJoCo physics simulator \cite{todorov2012mujoco}. We implemented our algorithm using tensorflow \cite{abadi2016tensorflow} and used openai gym \cite{openai} to interact with the Mujoco environment. In the MuJoCo environment, we teach a two-dimensional, one-legged robot called hopper, introduced in \cite{erez2012infinite}, to perform a backflip. The state representation of hopper is given by an $11$-dimensional real-valued vector $(S \in R^{11})$ and actions are a $3$ dimensional continuous vector $(A \in R^3)$. 

\par We teach the hopper robot to perform a backflip in the MuJoCo environment from a single video demonstration and human feedback. We use the video demonstration from another, pre-trained hopper agent performing a backflip to train our leaning agent. We start with a video of the same robot initially such that there is a one-to-one mapping between frames in the video and the standard representation. This reduces ambiguity and simplifies training and evaluation of the similarity predictor. In section \ref{sec:sim_pred}, we use this demonstration to test variations of the similarity predictor architecture. Then, we use the best performing variation of the similarity predictor and demonstrate its effectiveness in teaching the agent to backflip in section \ref{sec:syn_demo}. In section \ref{sec:real_demo}, we use a YouTube video of a human performing a backflip to teach the hopper robot how to backflip. The kinematics of the hopper agent and the human are quite different leading to ambiguity in mapping between visual and standard representations. We show that our method is effective even under the existence of these mapping ambiguities. Finally, in section \ref{sec:transfer}, we show that our similarity predictor, trained for backflipping, can be easily transferred to learn another task like hopping.

\subsection{Similarity Predictor}
\label{sec:sim_pred}
\par The goal of the similarity predictor network is to output a rating given a video clip and a list of state-action pairs indicating the similarity between their behaviors. The architecture of the similarity predictor is given in figure \ref{fig:similarity_network}. We perform a range of modifications to the similarity predictor network and evaluate them in this section. 

\par To test the similarity predictor network, we collect comparison ratings for 400 clips of the video demonstration and agent behavior pairs. We use the video demonstration shown in figure \ref{fig:syn_example}(a) of a pre-trained hopper agent performing a backflip. We randomly select $0.3$ second long clips for human annotation. The clips from the agent are generated by taking random actions similar to the pre-training stage of our algorithm described in section \ref{sec:arch}. The clips from agents behavior and expected video are shown to the human to obtain a rating from $1$ to $5$ indicating how similar the behaviors are.

\begin{table}[htb]
    \centering
    \resizebox{0.92\columnwidth}{!}{
    \begin{tabular}{|l|l|l|l|l|}
    \hline
                           & Val Acc & Test Acc & $F_1$-3,4,5 & $F_1$-4,5 \\ \hline
    Random sampling            & 54.22               & 42.89         & 13.46                       & 6.97                          \\ \hline
    Sampling equally           & 47.66               & 39.28         & 13.64                       & 18.81                         \\ \hline
    Class weights        & 52.05               & 38.11         & 14.14                       & 5.26                          \\ \hline
    Equal+Weights  & 49.05               & 38.16         & 12.79                       & 4.66                          \\ \hline
    Additional layer & 52.45               & 41.67         & 17.42                       & 22.68                         \\ \hline
    \end{tabular}
    }
    \caption{Validation and test accuracies of several variants of training of similarity network. Table also shows the $F_1$ score for predicting classes $3,4,5$ and classes $4,5$.}
    \label{tab:accuracy}
\end{table}

\par We split the annotated data into 3 parts. We use 200 annotations for training, 100 for validation and 100 for testing. As the learning agent's behavior is initially generated by taking random actions, the collected dataset is highly skewed. A large number of comparisons have a rating of $1$ with the number of comparisons with higher rating progressively decreasing (i.e., most of the random actions were very dissimilar to the behavior depicted in the demonstration video and thus received a low rating from the human). As our agent learns to take actions that result in a higher rating, it is important to minimize the errors in rating prediction (especially errors with higher ratings as these are more rare). To counteract this rating imbalance, we test several class-imbalance learning techniques to modify the training of the similarity network to work with the skewed dataset.  Table \ref{tab:accuracy} shows the validation and test accuracies for each of the training modifications of the similarity network. It also shows $F_1$ score for classes $3,4$ and $5$ together and for classes $4$ and $5$ together. All the accuracies and $F_1$ scores in table \ref{tab:accuracy} are averaged over $2$ runs with the similarity network initialized with random weights. 

\par Our first approach \textbf{(Random sampling)} is trained by sampling each training instance uniformly at random from the training data. As can be observed from table \ref{tab:accuracy}, although the overall validation and test accuracy of this method is relatively high, the $F_1$ score for class labels $3,4,5$ is low due to the skewed dataset. Our next approach was to train using batches containing equal numbers of instances from each class \textbf{(Sampling equally)}, where the instances of each class in a batch are randomly sampled and in equal proportions. We can notice from Table \ref{tab:accuracy}, the $F_1$ score for classes $4$ and $5$ increases, but the overall accuracy is slightly decreased. Another solution is to increase the weight of the cross entropy loss for misclassifying class labels with fewer instances \textbf{(Class weights)}. Surprisingly, this did not lead to an increase in accuracy or an increase in $F_1$ score. Next, we combined both of these solutions \textbf{(Equal + Weights)}, however the accuracy and $F_1$ score did not increase. Overall, the method in which we sample each class equally performs best for classes $3,4$ and $5$. Finally, we modified the architecture shown in figure \ref{fig:similarity_network} by adding a fully connected layer before the softmax layer and sampling equally from each class \textbf{(Additional layer)}. This network had the highest $F_1$ score for classes $3,4$ and $5$ without significant loss in overall validation and test accuracy. Also, most of the errors by this method were ``near misses''. Only $3.56\%$ of samples had an absolute error of $4$ and $4.56\%$ had an absolute error of $3$, where the absolute error is an absolute value of the difference between predicted and ground truth ratings. An absolute error of zero indicates correct prediction, while an absolute error of 1 indicates a prediction off by 1 (i.e. prediction of 3 with a true label of 4) and so on. Given the performance of additional layer method, in the rest of the paper we train with batches containing an equal number of samples from each class and with an additional fully connected layer. 

\subsection{Evaluation of learning from observations}
\label{sec:syn_demo}

\par In this section, we use the optimized similarity predictor network from section \ref{sec:sim_pred} to teach a reinforcement learning agent to backflip from a video demonstration. We use the rating from the similarity network as the reward for the reinforcement learning agent. The RL agent uses this reward to generate new trajectories that are similar in behavior to the video demonstration. We extract segments of these trajectories and corresponding frames from the demonstration. These clips are then annotated by the user asynchronously. Thus, further training of the similarity network, annotation and reinforcement learning are performed asynchronously. We show the effectiveness of our method in performing a backflip in the MuJoCo simulator. We also compare our method to a traditional reinforcement learning algorithm using a hand-coded reward function as well as the learning from human preferences method \cite{Christiano}.

\par
\begin{figure}[htb]
\begin{center}
\includegraphics[width=\columnwidth]{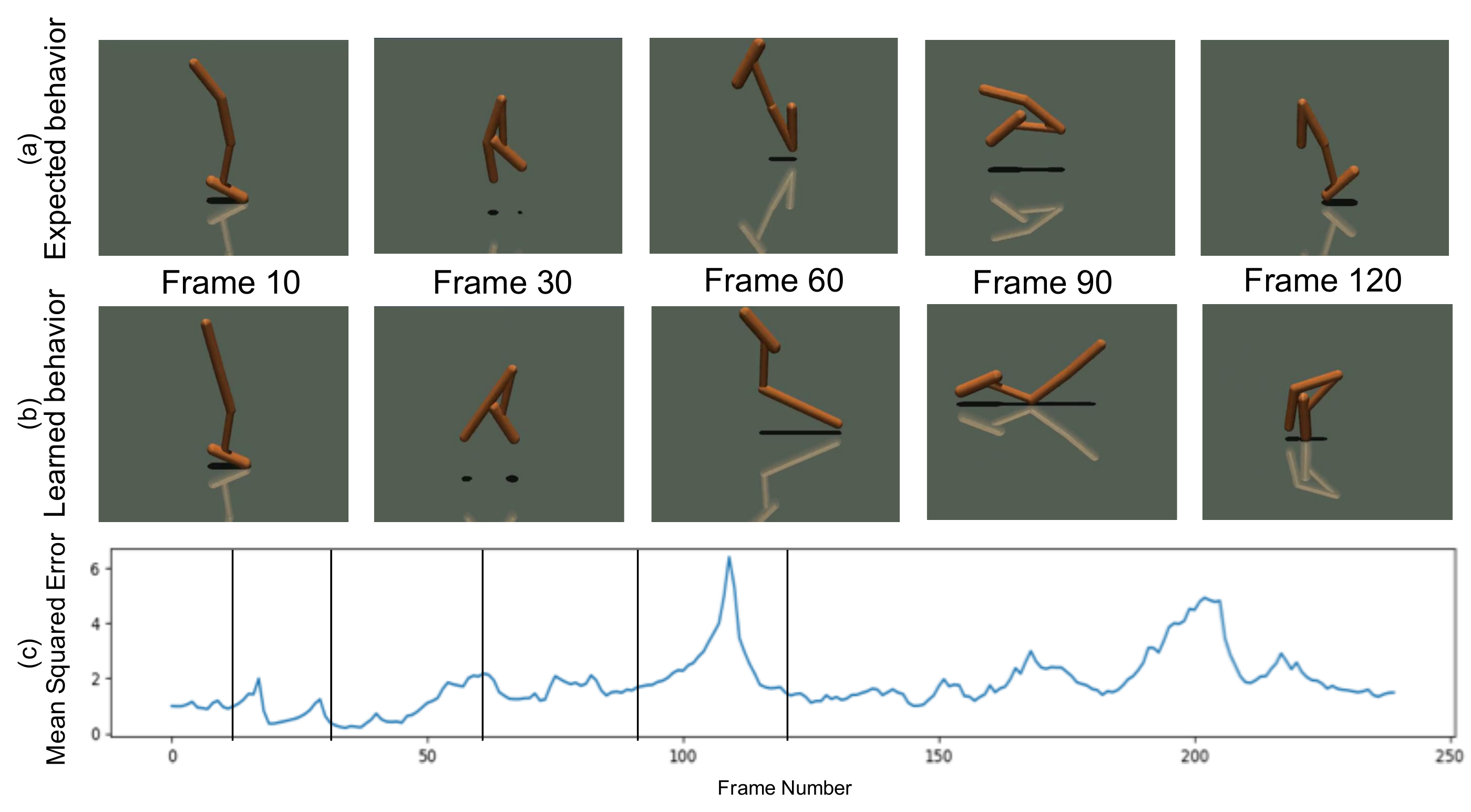}
\end{center}
\caption{Comparison of the backflip in the video demonstration and by our agent. (a) Frames from input video demonstration, (b) backflip performed by our agent and (c) mean squared error between state representation of trajectories in input video demonstration and by our agent}
\label{fig:syn_example}
\end{figure}

 For the traditional reinforcement learning comparison, we used a reward function from \cite{Christiano} that was constructed to get a hopper agent to backflip, where the more backflips the agent achieves in a given period of time, the higher reward it receives. We use this ``backflip reward function'' to train a traditional reinforcement learning agent using trust region policy optimization (TRPO)\cite{frans2016parallel}. We save the video and corresponding standard representation of the trajectory of the TRPO agent with the maximum backflip reward. The video is $8$ seconds long and its frames are shown in figure \ref{fig:syn_example}(a). We use this video as the demonstration video for teaching our agent to replicate the backflip. This enables us to track the progress of our agent using the backflip reward function.

\par We also compare our method to \textbf{learning from human preferences} by \cite{Christiano} as it also teaches autonomous agents new skills by learning from human feedback and does not use predefined reward function. \cite{Christiano} learns the reward function by collecting human preferences between pairs of trajectory segments. These annotations are used to train a reward predictor such that the reward for the preferred trajectory is maximized using cross entropy loss. Readers can refer to \cite{Christiano} for more details on the architecture. To learn from human preferences, we collect preferences with the goal of performing as many backflips as possible in $8$ seconds.

\par For both our method and \cite{Christiano}, we collected $200$ annotations during a pretraining stage where trajectory segments are extracted from rollouts of a random policy. We then collect $150$ annotations using the rollouts from the policy network that is being optimized asynchronously. Thus, we use 350 total annotations for training our method and learning from human preferences method.

\begin{table}[htb]
    \resizebox{0.95\columnwidth}{!}{
    \begin{tabular}{|l|l|l|l|}
    \hline
    Algorithm  & \makecell{\# of Human \\samples}  & \makecell{Maximum \\backflip \\ reward} & \makecell{Total \\backflips} \\ \hline
    TRPO                    & n/a          & 19859.81                  & 2                     \\ \hline
    \cite{Christiano} & 350 & 2064.27                   & 1                     \\ \hline
    Our method    & 350 & 3365.18                   & 2                     \\ \hline
    \end{tabular}
    }
    \caption{Performance comparison on the Backflipping task}
    \label{tab:env_reward}
\end{table}

\par We used two metrics to compare our method's performance: 1) total reward achieved from the backflip reward function and 2) total number of backflips performed during an episode (counted manually from the produced trajectory). Table \ref{tab:env_reward} shows the maximum backflip reward and corresponding number of backflips performed in $8$ seconds. Note that all three methods are solving different objective functions, using different inputs. The TRPO baseline uses a handcrafted reward function, our method uses video demonstration with human ratings and \cite{Christiano} uses human preferences between trajectories. Not surprisingly, the traditional TRPO agent achieves the highest reward as it was directly optimizing over the backflip reward function. Even still, our method, which does not require a reward function to optimize over, achieves higher reward compared to the learning from preferences method. Even though our method does not achieve the same amount of reward as the TRPO baseline, our agent performs same number of backflips as that of the TRPO baseline achieved and one more backflip than the preferences method achieved. We can get comparable performance to the TRPO baseline without requiring the use of a handcrafted reward function, and can achieve significantly better performance compared to the human preference method using the same amount of human feedback. Additionally, our method has the potential to learn any task by utilizing the vast number of videos on the web.

\par Figure \ref{fig:syn_example}(a) shows few frames from the video clip of the trajectory with maximum backflip reward generated using the TRPO algorithm. Figure \ref{fig:syn_example}(b) shows the behavior generated using our method. Figure \ref{fig:syn_example}(c) shows the mean squared error between state representations of the trajectories shown in figures \ref{fig:syn_example}(a) and (b) with respect to frame number/timesteps. The frame number of the frames displayed in figure \ref{fig:syn_example}(a) and (b) are indicated by vertical lines. Qualitatively, we can observe that backflip is performed in both video demonstration and by our agent. Notice that at frame 90 the agent's behavior diverges from the demonstration, but they reach a similar position at frame 120 and 135. If the agent takes a few dissimilar actions, the mean-squared error between the two trajectories will increase over the next few timesteps until it finds a state similar to the demonstration. After reaching the similar state, our agent again synchronizes with the video demonstration. The small periods of divergence between our agent's trajectory and the demonstrated trajectory typically occur if video demonstration contains successive rapid movements. This is the reason we see a spike in the mean-squared error at frame 110. Nevertheless, our agent successfully returns to a similar state at the end and replicates the video demonstration.

\subsection{Case study: Learning from a YouTube Video}
\label{sec:real_demo}

\begin{figure}[htb]
\begin{center}
\includegraphics[width=0.95\columnwidth]{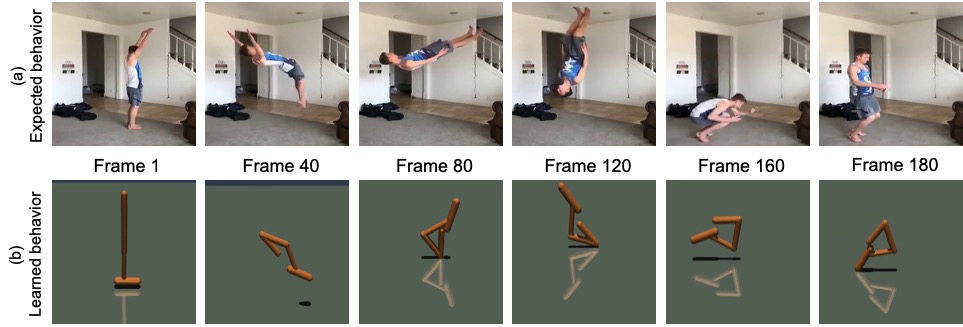}
\end{center}
\caption{Comparison of the backflip in the real-world video demonstration and by our agent. (a) Frames of YouTube video of
person performing backflip, (b) backflip performed by our agent }
\label{fig:real_example}
\end{figure}


\par In this section we demonstrate the ability of our method to learn from an actual YouTube video. We use the optimized similarity predictor to teach our hopper agent to backflip using a real-world video from YouTube of a human performing a backflip. The frames of the video are shown in figure \ref{fig:real_example}(a). It can be easily observed from Figure~\ref{fig:real_example} that since the kinematics of the human and hopper robot are very different, the mapping between between them is ambiguous. This makes both the annotation and learning of the behavior a challenging problem. The video demonstration is $6$ seconds long containing 180 frames. Using our method, we collected $200$ annotations during a pre-training stage and $165$ annotations while the reinforcement learning policy was being optimized. We used the same architecture and hyperparameters for training as was described previously in section \ref{sec:syn_demo}. Figure \ref{fig:real_example}(b) shows the learned backflip performed by the RL agent trained using our method. This shows the effectiveness of our method in replicating behavior demonstrated in single, non-annotated YouTube video.

\subsection{Case study: Transfer learning to hopping task}
\label{sec:transfer}

\begin{figure}[htb]
\begin{center}
\includegraphics[width=0.92\columnwidth]{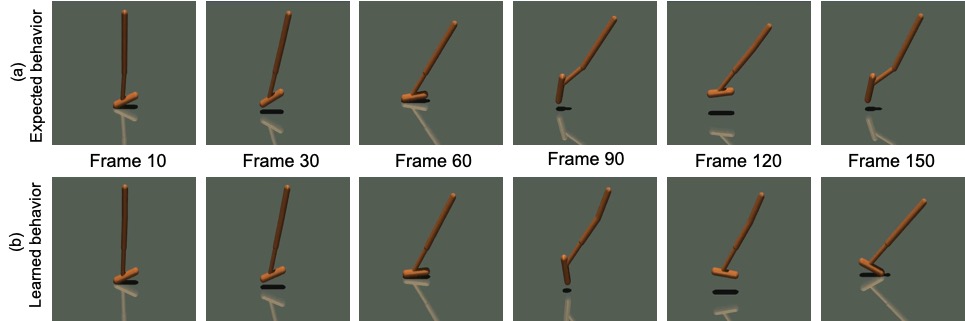}
\end{center}
\caption{Comparison of the hopping in the video demonstration and by our agent. (a) Frames from input video demonstration, (b) hopping imitation performed by our agent }
\label{fig:transfer_example}
\end{figure}

In this section, we test the generalization of our method to new tasks. Specifically, we test how well our trained similarity predictor network can transfer to a new task such as hopping. Although both tasks involve a jumping action, they require vastly different kinematic motions. To tackle this transfer learning problem, we fine-tuned our similarity predictor which was previously trained on the backflipping task. For fine-tuning, we collected new similarity feedback samples from the human and used them to update the network weights. Our hypothesis is that if our similarity prediction network has any capacity for generalization, then it will have higher reward for the hopping task with the same number of human samples when transferring from the backflip task to the hopping task, indicating that it learned a mapping between standard and visual representations that is not overfit to a single task.

\begin{table}[htb]
    \resizebox{\columnwidth}{!}{
    \begin{tabular}{|l|l|l|}
    \hline
    Algorithm  & \makecell{\# of Human \\samples}  & \makecell{Maximum \\hopping reward}  \\ \hline
    TRPO                    		& n/a          	& 1006.89                             \\ \hline
    Fine-tuning                   	& 55 			& 767.16                              \\ \hline
    Trained from scratch        		& 55         	& 433.28                           \\ \hline
    \end{tabular}}
    \caption{Transfer learning performance on the hopping task Note, only our method was transfered. The TRPO result shown is trained from scratch.}
    \label{tab:transfer_reward}
\end{table}

\par We tested the ability of our method to transfer to the hopping task by fine-tuning the similarity predictor and comparing it to a similarity predictor trained from scratch. Table \ref{tab:transfer_reward} shows the transfer learning results for the hopping task compared to the baseline TRPO method. We collected $55$ human annotations, with $30$ collected during the pre-training stage and $25$ while optimizing the RL agent. The $30$ annotations collected during the pre-training stage were used to train a similarity predictor initialized with random weights as well as the network initialized with transferred weights.The agent trained with the finetuned similarity predictor achieves $767.16$ reward while the agent trained with the similarity predictor initialized from scratch achieves $433.28$ maximum hopping reward. Figure \ref{fig:transfer_example} shows the qualitative comparison of the agent hopping in video demonstration and by our agent. Figure \ref{fig:transfer_example}(a) shows few frames from the video clip of the trajectory with the maximum hopping reward generated using the TRPO algorithm (the demonstration video). Figure \ref{fig:transfer_example}(b) shows the behavior generated using our method. Qualitatively, we can observe that the hopping task performed in both the video demonstration and by our agent are quite similar.A key takeaway here is that on average, we get significantly higher reward using a small amount of human data to learn a new task (only 55 samples) than if we had trained our method from scratch.

\section{Conclusion and Future work}
\label{sec:discussion}

\par We presented a framework for teaching new skills to autonomous agents from a single video demonstration using human feedback. In doing so, we leverage the advantages of both  the standard representation and visual representation. We train the reinforcement learning policies of robotic agents in their standard representation but leverage the rich visual representation for providing demonstrations of the desired behavior. Our method achieves this by learning a mapping between these two representations through the use of human feedback in the form of a similarity rating. We proposed several variations for constructing our similarity predictor function and compared their effect on its ability to model human similarity ratings. Using this learned predictor function, our method can imitate complex tasks like performing backflips using a video demonstration of a hopper agent backflipping as well as real-world video clip of a human backflipping taken from YouTube. Finally, we illustrated our methods ability to generalize to new tasks using transfer learning to quickly fine-tune our method to learn a hopping task.

\par The  limitation of current interface is that the user cannot explicitly select individual segments of the demonstration video where the agent has made a mistake. In future work, we would allow the user to select part of clip where the agent has made a mistake. This would allow user to give more granular feedback thus allowing it to imitate even more complex tasks. Nevertheless, we believe this work pushes the boundaries on creating easy, interactive ways for teaching new skills to autonomous agents and opens interesting directions to explore in the future.

\bibliographystyle{named}
\bibliography{bibliography}

\end{document}